\documentclass{article}
\usepackage{spconf,amsmath,graphicx}
\usepackage{import}
\usepackage[utf8]{inputenc}
\usepackage{adjustbox}
\usepackage{pgfplots}
\pgfplotsset{width=7cm,compat=1.5}

\title{Perfect match: Improved cross-modal Embeddings for audio-visual synchronisation}

\name{Soo-Whan Chung$^{1,2}$, Joon Son Chung$^{2}$ and Hong-Goo Kang$^{1}$}
\address{$^{1}$Department of Electrical \& Electronic Engineering, Yonsei University, Seoul, South Korea\\ $^{2}$Naver Corp., Seongnam-si, Gyeonggi-do, South Korea}

\begin{document}

\maketitle

\begin{abstract}
This paper proposes a new strategy for learning powerful cross-modal embeddings for audio-to-video synchronisation. Here, we set up the problem as one of cross-modal retrieval, where the objective is to find the most relevant audio segment given a short video clip. The method builds on the recent advances in learning representations from cross-modal self-supervision.
The main contributions of this paper are as follows: (1) we propose a new learning strategy where the embeddings are learnt via a multi-way matching problem, as opposed to a binary classification (matching or non-matching) problem as proposed by recent papers; (2) we demonstrate that performance of this method far exceeds the existing baselines on the synchronisation task; (3) we use the learnt embeddings for visual speech recognition in self-supervision, and show that the performance matches the representations learnt end-to-end in a fully-supervised manner.

\end{abstract}


\section{Introduction}
\label{sec:intro}
There has been a growing amount of interest in self-supervised learning, which has a significant advantage over fully supervised methods that has been prevalent over the past years, in that one can capitalize on the huge amount of data freely available on the Internet without the need for manual annotations. 

One of the earlier adaptations of such idea is the work on auto-encoders~\cite{hinton2006reducing}; there are recent work on learning representations via data imputation such as predicting context by inpainting~\cite{pathak2016context} or RGB images from only grey-scale images~\cite{zhang2016colorful}.
Recently, the use of {\em cross-modal} self-supervision has proved particularly popular, where the supervision comes from the correspondence between two or more naturally co-occurring streams, such as sound and images. 

Previous work of particular relevance is~\cite{Chung16a}, which uses a convolution neural network (CNN) model called SyncNet to learn a joint embedding of face image sequence and corresponding audio signal for lip synchronisation. The method learns powerful audio-visual features that are effective for active speaker detection and lip reading. \cite{halperin2018dynamic} has shown that the SyncNet features can also be used to achieve a dynamic temporal alignment between speech and video sequences for synchronising re-recorded speech segments to a pre-recorded video.

More recent (concurrent) papers have proposed methods for co-training audio and video representations using two-stream architectures for source localization~\cite{Arandjelovic18,Senocak18}, cross-modal retrieval~\cite{Arandjelovic18}, AV synchronisation and action recognition~\cite{Korbar18,Owens18} in videos of general domain. All these train two-stream networks to predict whether the audio and the video inputs are matching or not. The models are trained either with contrastive loss~\cite{Korbar18} or as a binary classification~\cite{Arandjelovic18,Senocak18,Owens18}. Similar strategies have also been used for cross-modal biometric matching between faces and voices~\cite{Nagrani18,kim2018learning}. Although these works show great promise on the cross-modal learning task, the question remains as whether these objectives are suitable for the proposed applications such as recognition and retrieval.

In this paper, we propose a novel training strategy for cross-modal learning, where we learn powerful cross-modal embeddings through a multi-way matching task. In particular, we combine the similarity-based methods ({\em e.g.} L2 distance loss) used to learn joint embeddings across modalities, with a multi-class cross-entropy loss; this way, the training objective naturally lends itself to cross-modal retrieval, where the task is to find the {\em most} relevant sample in one domain to a query in another modality. We propose a new training strategy in which the network is trained for the multi-way matching task without explicit class labels, whilst still benefiting from the favourable learning characteristics of the cross-entropy loss.

The effectiveness of this solution is demonstrated for audio-visual synchronisation, where the objective is to locate the most relevant audio segment given a short video clip. The models trained for multi-way matching is able to produce powerful representations of the auditory and visual information that can be applied to other tasks -- we also demonstrate that the learnt embeddings show better performance on a visual speech recognition task compared to the representations learnt via pairwise objectives.

\section{Architecture and training}
\label{sec:methods}

In this section, we describe the architecture and training strategy for the audio-visual matching task, and compare it to the existing state-of-the-art methods for audio-visual correspondence, including AVE-Net~\cite{Arandjelovic18} and SyncNet~\cite{Chung16a}.

\begin{figure}[t]
\begin{minipage}[b]{0.49\linewidth}
  \centering
  \centerline{\includegraphics[width=1\linewidth]{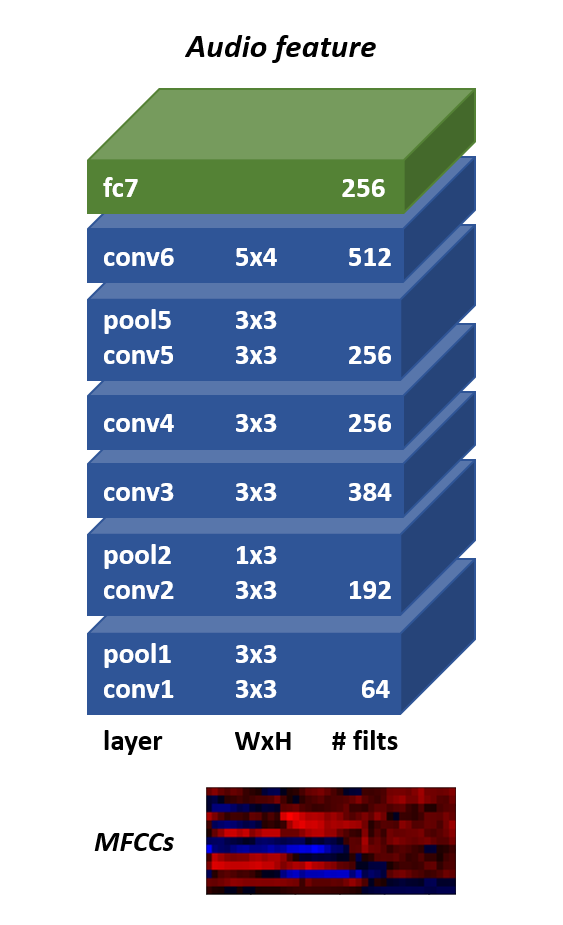}}
  \centerline{(a) Audio stream}\medskip
\end{minipage}
\begin{minipage}[b]{0.49\linewidth}
  \centering
  \centerline{\includegraphics[width=1\linewidth]{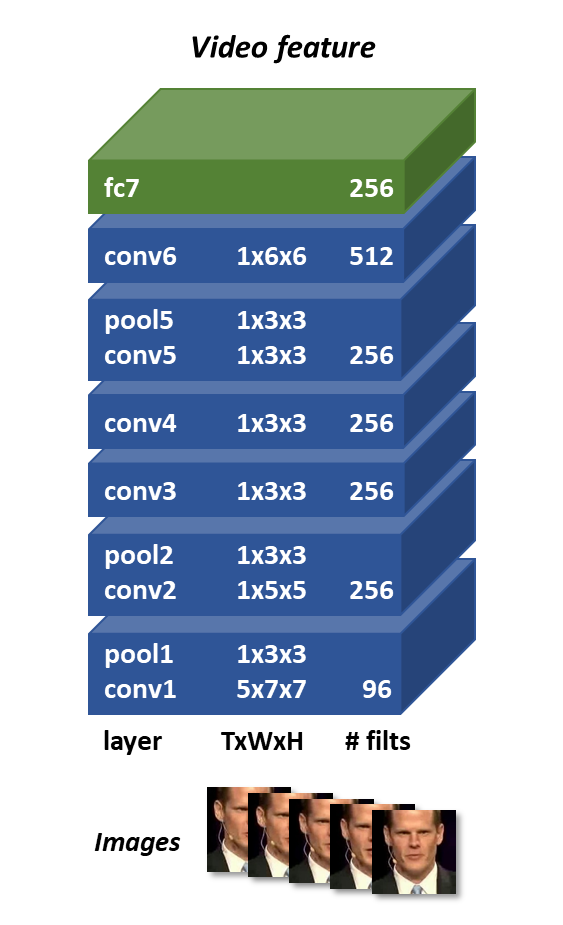}}
  \centerline{(b) Visual stream}\medskip
\end{minipage}
\caption{Trunk architecture for audio and visual stream}
\label{fig:streams}
\end{figure}

\begin{figure*}[t]
\centering
\begin{minipage}[b]{0.6\columnwidth}
  \centering
  \centerline{\includegraphics[width=5.0cm]{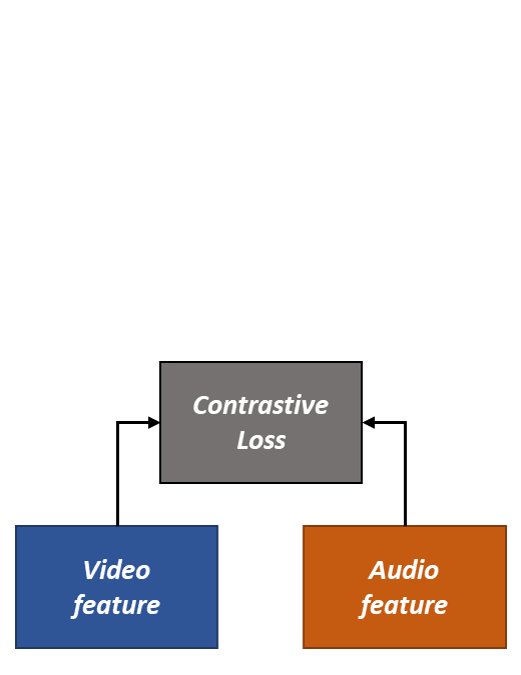}}
  \centerline{(a) SyncNet}\medskip
\end{minipage}
\begin{minipage}[b]{0.6\columnwidth}
  \centering
  \centerline{\includegraphics[width=5.0cm]{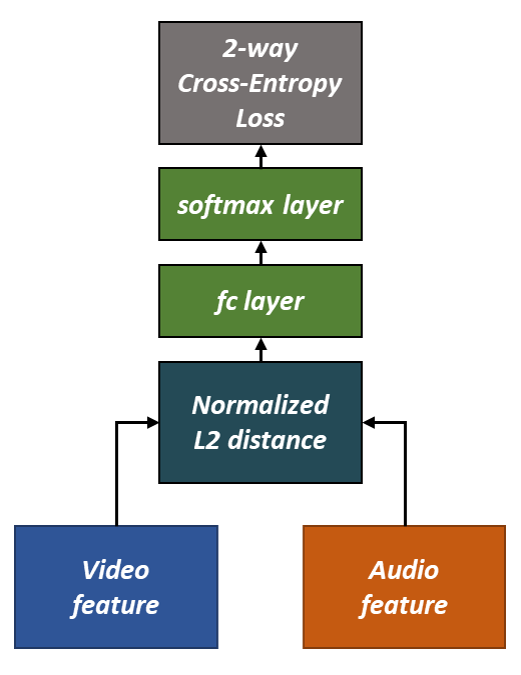}}
  \centerline{(b) AVE-Net}\medskip
\end{minipage}
\begin{minipage}[b]{0.6\columnwidth}
  \centering
  \centerline{\includegraphics[width=5.0cm]{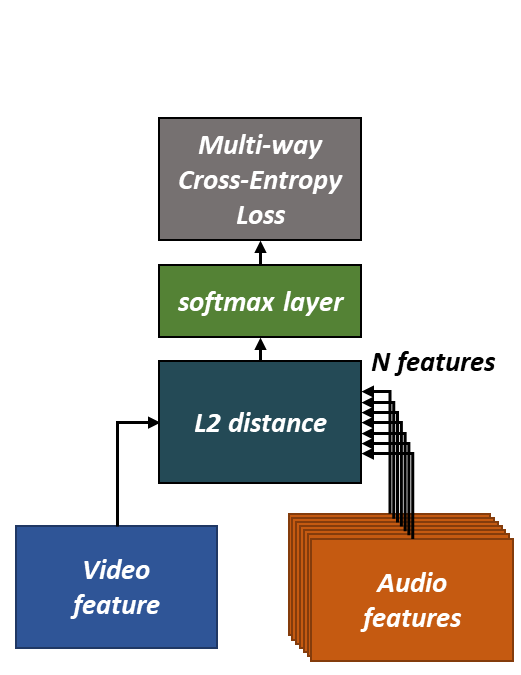}}
  \centerline{(c) Proposed model}\medskip
\end{minipage}
\caption{Comparison between the existing and proposed training strategies.}
\label{fig:ss}
\end{figure*}

\subsection{Trunk architecture}
\label{subsec:trunkarc}

The architecture of the audio and the video streams is described in this section. The inputs and the layer configurations are the same as SyncNet~\cite{Chung16a}, so that the performance using the new training strategy can be compared to the existing methods. The network ingests 0.2-second clips of both audio and video inputs.\vspace{5pt} \newline
\noindent\textbf{Audio stream.}
The inputs to the audio stream are the 13-dimensional Mel-frequency cepstral coefficients (MFCCs), extracted at every 10ms with 25ms frame length. Since the audio data is extracted from the video, there are natural environmental factors such as background noise and distortions in speech. The input size is 20 frames in the time-direction, and 13 cepstral coefficients in the other direction (so the input image is $13\times20$ pixels). The network is based on the VGG-M~\cite{Chatfield14} CNN model, but the filter sizes are modified for the audio input size as shown in Figure~\ref{fig:streams}(a).\vspace{5pt} \newline
\noindent\textbf{Visual stream.}
The input to the visual stream is a video of a cropped face, with a resolution of 224$\times$224 and a frame rate of 25 fps. The network ingests 5 stacked RGB frames at once, containing the visual information over the 0.2-second time frame. The visual stream is also based on the VGG-M~\cite{Chatfield14}, but the first layer has a filter size of $5\times7\times7$ instead of $7\times7$ of the regular VGG-M, in order to capture the motion information over the 5 frames. The detailed visual stream is described in Figure~\ref{fig:streams}(b).

\begin{figure}[ht]
  \centering
  \centerline{\includegraphics[width=1\linewidth]{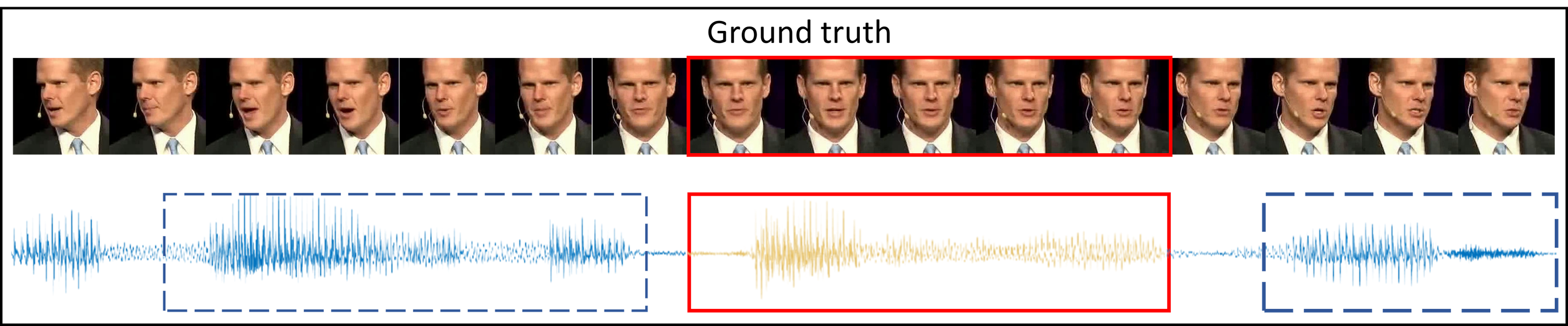}}
\caption{Sampling strategy for self-supervised learning. The red rectangle highlights the audio segments that corresponds to the talking face above, the blue dotted rectangles show non-matching audio segments.}
\label{fig:arc_sampst}
\end{figure}

\subsection{Training strategies}
\label{subsec:training}

The objective is to learn cross-modal embeddings of the audio and the visual information using self-supervision. The two baselines are trained as a pair-wise correspondence task, whereas the proposed method is set up as a multi-way matching task.\vspace{5pt} \newline
\noindent\textbf{Baseline - SyncNet.}
The original SyncNet~\cite{Chung16a} is trained with a contrastive loss, which is designed to maximise the distance between features for non-matching pairs of inputs, and minimise the distance for matching pairs. The audio and the video for non-matching pairs are sampled from the same face track, but from different points in time. The method requires manual tuning of the margin hyper-parameter.\vspace{5pt} \newline
\noindent\textbf{Baseline - AVE-Net.}
The Audio-Visual Embedding Network (AVE-Net)~\cite{Arandjelovic18}, designed for cross-modal retrieval, also takes the outputs from the audio and the video networks as inputs. The input vectors are L2 normalised, then the Euclidean distance between the two normalised embeddings are computed, before being passed through a fully-connected layer and a softmax layer. The fully-connected layer essentially learns the threshold on the distance above which the features are deemed not to correspond.\vspace{5pt} \newline
\noindent\textbf{Proposed - Multi-way classification.} 
Unlike previous methods that use pairwise losses, the proposed embeddings are learnt here via a multi-way matching task. Since pairwise losses are only used for the binary matching, they do not use context information. However, the multi-way matching strategy controls not only the distance between pairs but also uses relevant information among sequential data to train the model. The learning criterion takes one input feature from the visual stream and multiple features from the audio stream. This can be set up as any $N$-way feature matching task. Euclidean distances between the audio and video features are computed, resulting in $N$ distances. The network is then trained with a cross-entropy loss on the {\em inverse} of this distance after passing through a softmax layer, so that the similarity between matching pairs are greater than that of non-matching pairs. Training strategies described here are summarised in Figure~\ref{fig:ss}.

All $N$ audio frames are sampled from the same face track as the video clip, but only one corresponds to the video clip in time. This is to force the network to learn the content of what is being said, rather than the identity or other utterance characteristics. The sampling strategy is illustrated in Figure~\ref{fig:arc_sampst}. 

\section{Experiments}
\label{sec:experiments}

In this section, we compare the performance of the proposed system to existing method for lip synchronisation and a related audio-visual application.

\subsection{Audio-to-video synchronisation}
\label{subsec:lipsync}
Audio-to-video synchronisation can be seen as a cross-modal retrieval task, where the temporal offset is found by selecting an audio segment from a set, given a video segment. This is done by computing the distance between a learnt video feature (from a 5-frame window) and a set of audio features. We assume that the two streams are synchronised when the distances between features are minimised. However as \cite{Chung16a} suggests, one visual feature might not be enough to determine the correct offset, since not all samples contain discriminative information -- for instance, there may be some 5-frame video segments in which nothing is said. Therefore, we also conduct experiments with the context window of more than 5 video frames, in which case we average the distances across multiple video samples (with a temporal stride of 1 frame).\vspace{5pt} \newline
\begin{figure}[t]
    \centering
    \begin{tikzpicture}
    \pgfplotsset{set layers}
    \begin{axis}[
        scale only axis,
        width=0.7*\linewidth,height=\linewidth*0.5,
        xlabel={The number of audio features ($N$)},
        ylabel={Accuracy (\%)},
        ymin=84, ymax=92,
        xtick={2,5,10,20,30,40,50,60},
        xticklabels={2,5,10,20,30,40,50,60},
        ytick={84,85,86,87,88,89,90,91,92},
        ymajorgrids=true,
        legend columns=1,
        legend style={at={(0.5,-0.3)},anchor=north},
        grid style=dashed
    ]
    \addplot+[
        color=black,
        mark=square,
        ]
        coordinates {
        (2,85.2)(5,87.2)(10,87.8)(20,87.8)(30,88.2)(40,89.6)(50,87.1)(60,86.8)
        };\label{plot_one}
        \addlegendentry{Accuracy (\%)}
    \end{axis}
    \begin{axis}[
        scale only axis,
        width=0.7*\linewidth,height=\linewidth*0.5,
        ylabel={\# of video clips},
        axis y line*=right,
        axis x line=none,
        ymin=0, ymax=100000,
        xtick={2,5,10,20,30,40,50,60},
        xticklabels={2,5,10,20,30,40,50,60},
        ytick={0,20000,40000,60000,80000,100000},
        legend columns=2,
        legend style={at={(0.5,-0.3)},anchor=north},
        grid style=dashed   
    ]
    \addlegendimage{/pgfplots/refstyle=plot_one}\addlegendentry{Accuracy (\%)}
    \addplot+[
        color=black,
        mark=o,
        ]
        coordinates {
        (2,94826)(5,94768)(10,93975)(20,66725)(30,38033)(40,23942)(50,15697)(60,10784)
        };\label{plot_two}
    \addlegendentry{\# of video clips}
    \end{axis}
    \end{tikzpicture}
    \caption{Synchronisation accuracy according to {\em N}}
 \label{fig:acc_N}
\end{figure}
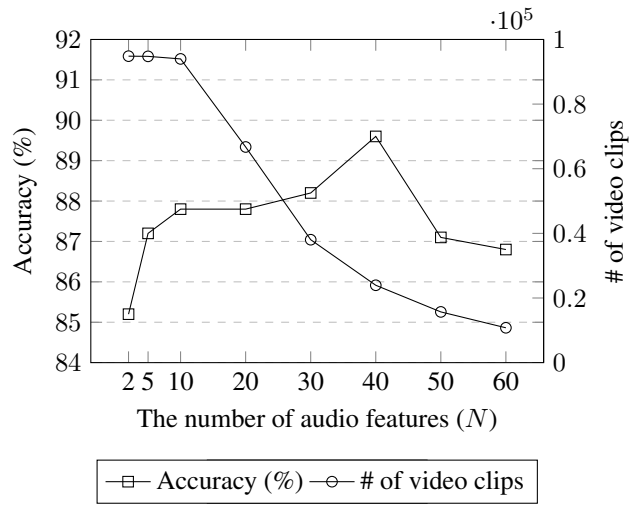
\noindent\textbf{Dataset.} 
The network is trained on the pre-train set of the Lip Reading Sentences 2 (LRS2)~\cite{Chung17} dataset. The LRS2 dataset contains 96,318 clips for training, and 1,243 for test. There is a trade-off between the number of classes (or candidate audio features) $N$ and the number of available video clips for training, since longer video clips are required to train networks with larger $N$ (the candidate audio clips are sampled without overlap). We run experiments with different values of $N$ in order to find the optimal value, and report the accuracy and the number of available video clips in Figure~\ref{fig:acc_N}.\vspace{5pt} \newline
\noindent\textbf{Evaluation protocol.}
The task is to determine the correct synchronisation within a $\pm$15 frame window, and the synchronisation is determined to be correct if the predicted offset is within 1 video frame of the ground truth. A random prediction would therefore yield 9.7\% accuracy. Since there are non-informative frames, we also compute the sync offset over various numbers of input visual frames, using the average distances between features for input length $K>5$. \vspace{5pt} \newline
\begin{table}[t]
\centering
\caption{Synchronization accuracy. {\bf \# Frames}: the number of visual frames for which the distances are averaged over.}
 \begin{tabular}{| r | r | r | r | } 
 \hline
 {\bf \# Frames} & SyncNet & AVE-Net & {\bf Proposed} \\ 
 \hline
 \hline
 5   & 75.8\%   & 74.1\%   & {\bf 89.5\%} \\ 
 7   & 82.3\%   & 80.4\%   & {\bf 92.1\%} \\ 
 9   & 87.6\%   & 86.1\%   & {\bf 94.7\%} \\ 
 11  & 91.8\%   & 90.6\%   & {\bf 96.1\%} \\ 
 13  & 94.5\%   & 93.7\%   & {\bf 97.5\%} \\ 
 15  & 96.1\%   & 95.5\%   & {\bf 98.1\%} \\ 
\hline
\end{tabular}
\normalsize
\label{table:result_sync}
\end{table}
\noindent\textbf{Results.}
The results of experiments are given using the network trained with $N=40$ in Table~\ref{table:result_sync}. The performance of the proposed method far exceeds the baseline trained with a pair-wise objectives. In particular, for \# frames = 5 ({\em i.e.} no context beyond the receptive field), there is a significant increase in synchronisation performance from 75.8\% to 89.5\%. 

\subsection{Visual speech recognition}
\label{subsec:lipreading}

\begin{figure}[t]
  \centering
  \centerline{\includegraphics[width=0.6\linewidth]{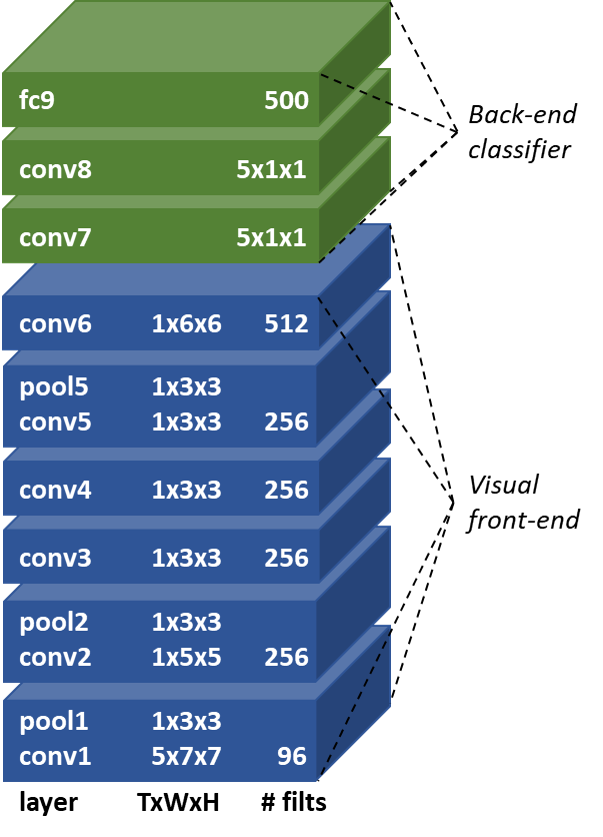}}
\caption{Architecture of the {\bf TC-5} lip reading network.}
\label{fig:arc_lr}
\end{figure}

The network learns a powerful embedding of the visual information contained in the input video. The objective of this experiment is to show that the embeddings learnt by the matching network are effective for other applications, in this case, visual speech recognition.  This is demonstrated on a word-level recognition task, and we compare the performance using the embeddings learnt by the proposed self-supervised method to networks trained end-to-end with full supervision.\vspace{5pt} \newline
\noindent\textbf{Dataset.} 
We train and evaluate the models on the Lip Reading in the Wild (LRW) \cite{Chung16b} dataset, which consists of word-level speech and video segments extracted from the British television. The dataset has a vocabulary size of 500, and contains over 500,000 utterances, of which 25,000 are reserved for testing.\vspace{5pt} \newline
\noindent\textbf{Architecture.} 
The front-end architecture is taken from the visual stream of the network described in Section~\ref{subsec:trunkarc}. We propose a 2-layer temporal convolution back-end, followed by a 500-way softmax classification layer. This network structure is summarised in Figure~\ref{fig:arc_lr} and is referred to as {\bf TC-5} in Table~\ref{table:result_lr}. The {\bf `5'} refers to the receptive field of the feature extractor in the temporal dimension, in line with the naming convention of the networks in~\cite{Chung18}. The performance of the {\bf TC-5} model exceeds the network designs proposed in~\cite{Chung18} when trained end-to-end (E2E). The visual features are extracted in advance for the `pre-trained' experiments (PT), and only the back-end layers are trained for the 500-way classification task -- the feature extractor is not fine-tuned with full supervision.\vspace{5pt} \newline
\noindent\textbf{Results.} 
\begin{table}[t]
\centering
\caption{Word accuracy of visual speech recognition using various architectures and training methods.}
 \begin{tabular}{| l | l | r | }
 \hline
 {\bf Architecture} & {\bf Method} & {\bf Top-1 (\%)} \\ 
 \hline
 \hline
 MT-5~\cite{Chung18} & E2E & 66.8   \\  \hline
 LF-5~\cite{Chung18} & E2E & 66.0   \\  \hline
 LSTM-5~\cite{Chung18} & E2E & 65.4 \\  \hline \hline
 TC-5 & E2E & 71.5  \\ \hline
 TC-5 & PT - SyncNet & 67.8  \\ \hline
 TC-5 & PT - AVE-Net & 66.7  \\ \hline
 TC-5 & PT - {\bf Proposed} & {\bf 71.6}  \\ \hline
\end{tabular}
\normalsize
\label{table:result_lr}
\end{table}
We report the results on the visual speech recognition task in Table~\ref{table:result_lr}. The results are compared to existing lip reading networks based on the VGG-M base architecture, and also to a model with the identical {\bf TC-5} architecture trained end-to-end on the large-scale LRW dataset in a full supervision. It is noteworthy that the performance of the feature extractor trained with the self-supervised method matches that of the end-to-end trained network without any fine-tuning.

\section{Conclusion}
\label{sec:conclusion}

We proposed a new training strategy for cross-modal matching and retrieval, which enables networks to be trained for matching without explicit class labels, whilst benefiting from favourable learning characteristics of the cross entropy loss. The experimental results show superior performance on the audio-visual synchronisation task compared to the existing state-of-the-art. The proposed embedding strategy also gives a significant improvement on the visual speech recognition task, and the performance matches that of a fully-supervised method with the same architecture. The method should also be applicable to other cross-modal tasks. \newline

\noindent\textbf{Acknowledgements.} 
We would like to thank Bong-Jin Lee, Dongyoon Han, Jaesung Huh, Min-Seok Choi and Youna Ji at Naver Clova AI for their helpful advice. 

\clearpage

\vfill
\newpage
\label{sec:refs}
\bibliographystyle{IEEEbib}
\bibliography{longstrings,refs}

\end{document}